\documentclass[10pt,twocolumn,letterpaper]{article}

\usepackage{cvpr}              % To produce the CAMERA-READY version
\usepackage{graphicx}
\usepackage{amsmath}
\usepackage{amssymb}
\usepackage{booktabs}
\usepackage{pifont}

\usepackage[pagebackref,breaklinks,colorlinks]{hyperref}

\usepackage[capitalize]{cleveref}
\crefname{section}{Sec.}{Secs.}
\Crefname{section}{Section}{Sections}
\Crefname{table}{Table}{Tables}
\crefname{table}{Tab.}{Tabs.}
\crefname{table}{Tab.}{Tabs.}

\usepackage{xcolor}
\usepackage{enumitem}
\usepackage{xspace}
\usepackage{booktabs}
\usepackage{colortbl}
\usepackage{multirow}
\usepackage{makecell}
\usepackage{lipsum} % Just for generating dummy text, can be removed
\usepackage{gensymb} % for \degree

\usepackage[labelsep=period]{caption}
\renewcommand{\paragraph}[1]{\vspace{0.2em}\noindent \textbf{#1 \hspace{0.2em}}}

\definecolor{MyDarkRed}{rgb}{0.66, 0.16, 0.16}
\definecolor{MyDarkBlue}{rgb}{0.16, 0.16, 0.66}

\def\cvprPaperID{7825} % *** Enter the CVPR Paper ID here
\def\confName{CVPR}
\def\confYear{2024}

\begin{document}

\title{Dynamic Policy-Driven Adaptive Multi-Instance Learning for Whole Slide Image Classification}

\author{Tingting~Zheng\textsuperscript{\rm 1}\quad Kui~Jiang\textsuperscript{\rm 1}\quad Hongxun~Yao\textsuperscript{\rm 1*}\\
\vspace{-1mm}
 {\textsuperscript{\rm 1} Harbin Institute of Technology}  \\ 
\vspace{-1mm}
{\tt\small 23b903051@stu.hit.edu.cn, \{jiangkui, h.yao\}@hit.edu.cn}\\ 
\vspace{-2mm}
{\tt\small \url{https://vilab.hit.edu.cn/projects/pamil}}
}

\maketitle

\begin{abstract}
Multi-Instance Learning (MIL) has shown impressive performance for histopathology whole slide image (WSI) analysis using bags or pseudo-bags. It involves instance sampling, feature representation, and decision-making. However, existing MIL-based technologies at least suffer from one or more of the following problems: 1) requiring high storage and intensive pre-processing for numerous instances (sampling); 2) potential over-fitting with limited knowledge to predict bag labels (feature representation); 3) pseudo-bag counts and prior biases affect model robustness and generalizability (decision-making). Inspired by clinical diagnostics, using the past sampling instances can facilitate the final WSI analysis, but it is barely explored in prior technologies. To break free these limitations, we integrate the dynamic instance sampling and reinforcement learning into a unified framework to improve the instance selection and feature aggregation, forming a novel Dynamic Policy Instance Selection (DPIS) scheme for better and more credible decision-making. Specifically, the measurement of feature distance and reward function are employed to boost continuous instance sampling. To alleviate the over-fitting, we explore the latent global relations among instances for more robust and discriminative feature representation while establishing reward and punishment mechanisms to correct biases in pseudo-bags using contrastive learning. These strategies form the final Dynamic Policy-Driven Adaptive Multi-Instance Learning (PAMIL) method for WSI tasks. Extensive experiments reveal that our PAMIL method outperforms the state-of-the-art by 3.8\% on CAMELYON16 and 4.4\% on TCGA lung cancer datasets. 
\end{abstract}

\renewcommand{\thefootnote}{}
\footnotetext{*Corresponding authors}

\section{Introduction}
\label{sec:intro}
\begin{figure}[tb] \centering
    \includegraphics[width=0.45\textwidth]{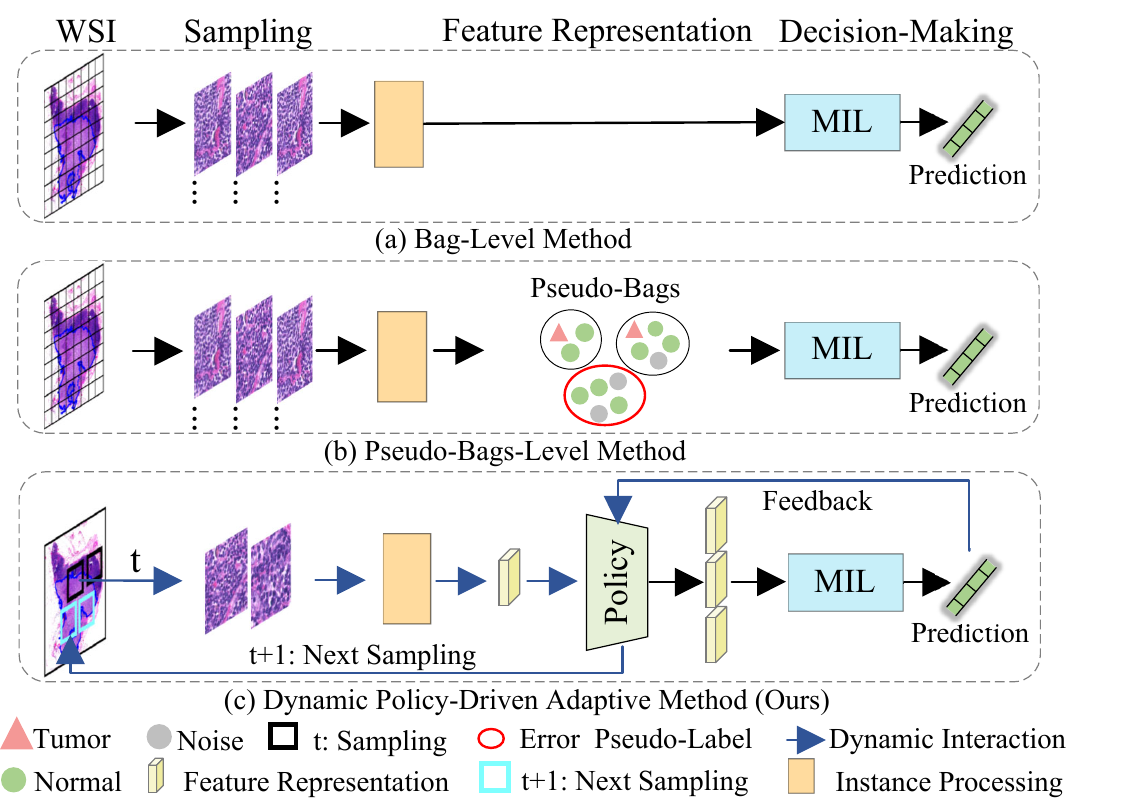}
\caption{Different schemes for Multi-Instance Learning (MIL). (a) Each bag is treated as a group. (b) Each bag is divided into pseudo-bags. (c) \textbf{The proposed Dynamic Policy-Driven Adaptive Multi-Instance Learning (PAMIL) Method}.} \vspace{-5mm} \label{fig:figure1}
\end{figure}

Histopathology slides provide detailed information on the morphology and scale of the tumor, being the gold standard for disease diagnosis~\cite{CAMELYON16,TCGALung}. However, complex biological structures in slides present a challenge for pathologists to discern intricate variances~\cite{CLAM,RNNMIL}. Recently, with the development of bioimaging and computation display technologies, digital whole slide images (WSIs) have emerged as effective diagnostic aids~\cite{ReviewMILWSI1,RLreview20}. In particular, a substantial progress has been achieved by introducing deep learning approaches for WSI analysis~\cite{DTFDMIL,MSDAMIL,HGNN_graph,CVPR21DSMIL,RankMix}. However, due to irregularities of tumor tissue and gigapixel resolution for WSIs, the collection of high-precision is laborious and time-consuming~\cite{CAMELYON16,TCGALung}. Weakly supervised solutions have drawn widespread attention than supervised schemes~\cite{Transmil,NCRF2018}. In particular, multi-instance learning (MIL)~\cite{MILframework} schemes emerge as the times require, providing alternative solutions for WSI analysis. Specifically, each WSI is described as a ``bag" containing numerous instances (or ``patches") sampled from the slide~\cite{ReviewMILWSI1}. A ``bag" is labeled with positive or negative tip, but its instances are label-free. The core mission of MIL-based methods is to aggregate information from all instances to provide bag-level predictions. According to grouping the instances, these methods can be broadly divided into two categories: bag-level and pseudo-bags-level methods, as illustrated in Figure~\ref{fig:figure1}. Bag-level-based methods pack all instances or embeds their features into the model to regress the final prediction~\cite{RNNMIL, ABMIL,CVPR21DSMIL,Transmil} (Figure~\ref{fig:figure1}(a)). Besides the computational and memory burden, it is non-trivial to identify the most informative features from thousands of patches~\cite{DTFDMIL,22pseudo-labelDgmil}. In particular, since the mutual relationships among instances are barely explored, the distribution bias between minor tumor regions and the extensive normal tissues exacerbate the risk of model over-fitting~\cite{IBMIL,ProtoDiv,MuRCL}.

By contrast, pseudo-bags-based methods leverage clustering or random schemes to decompose all instances or embedding features into multiple classes or groups, denoted as pseudo-bags (or sub-bags)~\cite{DTFDMIL,MMIL,MuRCL}. Then bag labels are assigned to the pseudo-bags as pseudo-labels, as shown in Figure~\ref{fig:figure1}(b)~\cite{IBMIL,ProtoDiv,Sub_Labels_Noisy,22pseudo-labelDgmil,RankMix}. With the guidance of pseudo labels, a substantial improvement has been achieved by pseudo-bags-based methods~\cite{DTFDMIL,Sub_Labels_Noisy,22pseudo-labelDgmil,RankMix}. But notably, the pre-processing for cluster and pseudo label generation is tedious. Besides that, there exist at least two other shortcomings: 1) The tumor instances may be scattered over wide groups, especially for WSIs with small and dispersed lesions, where the informative features are diminished. It consequently increases the risk of false negative reactions. 2) The pseudo-bags derived from random sampling disrupt intra-class instance homogeneity, while the pseudo-bags derived from clustering may overly prioritize contextual similarities irrelevant to the label. Both strategies heighten the risk of misalignment between pseudo-bags features and bag label, arising prediction errors~\cite{Sub_Labels_Noisy,ProtoDiv}.

To alleviate these issues, some efforts utilize attention mechanisms to adaptively focus on the label-relation components~\cite{DTFDMIL,ProtoDiv,RankMix}. However, capturing the most informative features from thousands of instances with a simple Sigmoid layer is a non-trivial issue. In addition, the feature relationships among continuous sampling groups are barely explored during the diagnostic process, and thus these methods are short of stability and robustness. This inspires us to ask the following question: \textit{Whether more elaborate and simplified sampling schemes and information aggregation mechanism can be developed to capture the most discriminative features for accurate prediction?}

By reviewing the procedures of existing MIL methods~\cite{CLAM,RNNMIL,MuRCL}, WSI analysis can be roughly decomposed into three sub-tasks: instance sampling, feature representation and decision-making (as shown in Figure~\ref{fig:figure1}). Although some efforts have highlighted that the current decision can benefit from the past knowledge and continuous learning~\cite{RLreview20,Time_Decision,RL3DSeg20CVPR,RL3DLoca19MIA,RLHer2,RL3DSeg19Miccal}, the latent priors among instance sampling, feature representation and decision-making are barely exploited. More specifically, the extracted features of current instances can encourage both sampling and feature representation of the next sampling stage to distill more informative components oriented towards the given label. However, introducing past information for guidance naturally raises two issues: \textit{1) How to sample the most informative instances from the rest of sampling bag; 2) How to aggregate historical features for robust decision making?}

To tackle the aforementioned issues, we improve the existing MIL by exploring the intrinsic relations of instance sampling, feature representation and decision-making, and advise a dynamic policy-driven adaptive multi-instance learning (PAMIL) framework for WSI classification (Figure~\ref{fig:figure1}(c)). To tackle the first question, unlike the continuous random sampling~\cite{MMIL}, we introduce dynamic instance sampling and the experience-based learning of reinforcement learning (RL)~\cite{RLMITbooks}, and construct the dynamic policy-driven instance selection (DPIS) scheme. Specifically, DPIS takes the local neighbor relation and distance similarity between the current instance representation and the remaining instances into consideration for guiding the instance selection. Additionally, a reward function aligned with the target and a self-guided punishment mechanism is introduced to encourage the model to focus on label-related and robust instances to obtain the most representative bag predictions. 

Besides the sampling optimization, how to aggregate the knowledge of continuous sampling stages to produce more distinguishable features is crucial for generating robust and accurate predictions. Since there exists uncertainty to drive the representative token from input instances, we thus propose a selection fusion feature representation (SFFR) strategy. Specifically, to achieve general and robust tokens, we introduce contrastive loss~\cite{CLsiamese} among the adjacent stage pseudo-bags to guide the generation of ``positive" features (related to the given label) in the current stage. Meanwhile, to avoid forgetting historical information, we fully exploit the capability of the class token (CLS) in Transformer~\cite{Transformer}, where the previous tokens serve as the initial representation of current stage to produce a more precise bag representation. Experimental results demonstrate that our proposed PAMIL framework significantly outperforms current state-of-the-art (SOTA) methods.

The main contributions of this paper are as follows:
\vspace{3pt}
\begin{itemize}[itemsep=0pt,parsep=0pt,topsep=2bp]
    \item We examine the sampling, representation, and decision-making processes in MIL tasks, and investigate their underlying connections to establish a dynamic policy-driven adaptive multi-instance learning framework (PAMIL) for precise bag-label inference.
    \item We pioneer a dynamic policy-driven instance selection (DPIS) method for sample selection. This is achieved by considering the local neighbor relationship and distance similarity between the current instance representation and the remaining instances.
    \item We advise a selection fusion feature representation (SFFR) method for more precise bag representation by fully exploiting the historical information of sub-bags.
\end{itemize}
\begin{figure*}[!htb] \centering
    \centering
    \includegraphics[width=0.9\textwidth]{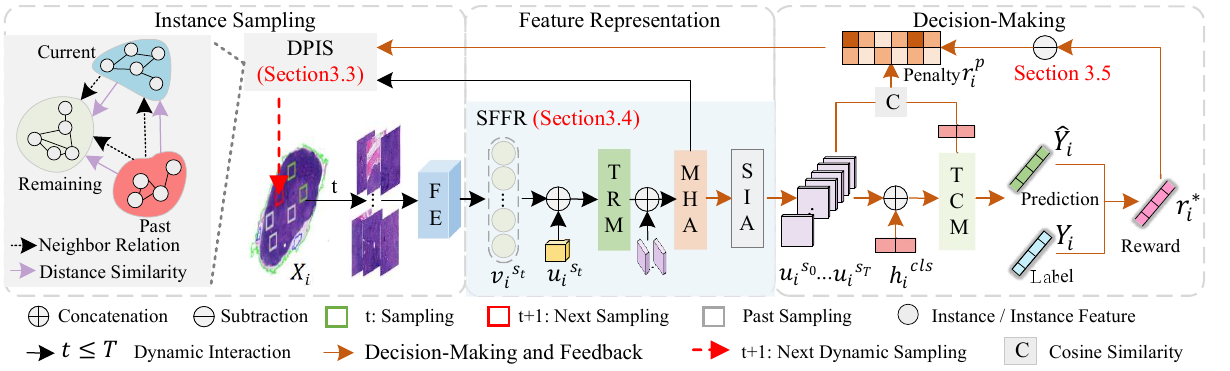}
    \caption{Architecture of our proposed dynamic policy-driven adaptive multi-instance learning (PAMIL) framework. PAMIL consists of a dynamic policy-driven instance selection (DPIS) scheme, a selection fusion feature representation (SFFR) module, and a Transformer-based classification module (TCM). DPIS learns relationships among past, current, and remaining instances of $X_i$, and samples instances embedded as $v_i^{s_t}$ using a feature extractor (FE). Then, SFFR takes $v_i^{s_t}$ and an initial token $u_i^{s_t}$ as inputs, refining $u_i^{s_t}$ by utilizing a Transformer module (TRM) and a multi-head attention (MHA) mechanism to fuse $v_i^{s_t}$ and past tokens. Meanwhile, we introduce a Siamese (SIA) structure between $u_i^{s_t}$ and $u_i^{s_{t-1}}$ to enhance the robustness of $u_i^{s_t}$. Finally, TCM uses a class token (CLS) $h_i^{cls}$ to aggregate $\{u_{i}^{s_t}\}_{t=1}^{T}$ for inferring the probability ${\hat{Y}_i}$ of $X_i$, providing feedback as reward $r_{i}^*$ and penalty $r_i^p$ to the DPIS.}
    \vspace{-5mm}
    \label{fig:figure2}
\end{figure*}

\section{Related Work}
\label{sec:related_works}
\subsection{Multi-Instance Learning in WSI Classification}%
MIL aims to determine bag labels from sampled instance representations. As shown in Figure~\ref{fig:figure1}, there are two primary MIL categories based on instance grouping: bag-level and pseudo-bag-level methods. The former applies diverse selection strategies, ranging from the conventional MaxPooling, MeanPooling, and Top-K~\cite{ReviewMILWSI1, RNNMIL, CLAM} to popular attention and Transformer~\cite{ABMIL, Transmil, CVPR21DSMIL, IAT}, to integrate most salient features for prediction. However, directly predicting the label from thousands of instances derived from gigapixel resolution WSIs often struggles to produce robust and satisfactory results. The latter tends to exploit correlations within instances or embedding features, employing clustering or random grouping to form pseudo-bags or labels to improve inference~\cite{MuRCL, MMIL, DTFDMIL,22pseudo-labelDgmil,ProtoDiv}. Nevertheless, grouping based on subtle correlations may disrupt intra-group feature consistency, leading to unstable predictions, especially for class-imbalanced datasets. To address this issue, some works use a teacher-student model to focus on hard instances~\cite{MHIM} or refine pseudo-bags features from classifier predictions~\cite{ RankMix,BackWSIfinetuing23,ReMix}. Besides depending on precise regression results, these methods still fall short in fundamentally ensuring that the sampled and aggregated representations accurately and consistently align with bag labels.

\subsection{Reinforcement Learning in Medical Imaging}%
Reinforcement learning (RL) has drawn increasing attention in computer vision tasks for handling long sequential data via dynamic interactions~\cite{RLRS, RLGPTCVPR23}. In particular, self-driven learning based on past experiences to correct biases for more accurate predictions is in line with clinical diagnosis~\cite{ReviewMILWSI1}. Some interesting practices have been applied in medical image landmark detection~\cite{RL3DLocal, RL3DLoca19MIA}, segmentation~\cite{RL3DSeg20CVPR, RL3DSeg19Miccal}, and classification~\cite{RLreview20,FastMDP,RLHer2}. 
To optimize the WSI tumor detection, few attention has been paid to explore historical information in MIL methods~\cite{FastMDP, RLHer2, RLogist}. Due to the absence of instance labels, the main challenge is to define rewards strategy. Moreover, subtle variations in designing state and action spaces can drastically affect model performance, with some not converging~\cite{RLreview20}. To tackle these challenges, we introduce a model-independent reward mechanism. In particular, an effective action-to-state mapping is achieved by employing the spatial correlations and feature similarity relations among instances.

\section{Proposed Method}%
\label{sec:method}

\subsection{Review for MIL Formulation}
\label{sec:MIL_Formulation}
For a binary classification task, given the dataset \(X = \{(X_{1},{Y_{1}}), (X_{2},{Y_{2}}), \ldots, (X_{N},{Y_{N}})\}\), it comprises \(N\) WSIs \(X_{i}\) and the corresponding labels \(Y_{i} \in \{0,1\}\). Taking $X_{i}$ as an example, it is treated as the $i^{th}$ ``bag" involving $B_i$ instances $x_{i,b}$ $\in \mathbb{R}^{W\times H \times 3}$ sampled from it, where $H$ and $W$ denote the height and width and $b\in[1, B_i]$. All instances $x_{i,b}$ are label-free, but part of the ``bag" label of \(Y_{i} \in \{0,1\}\). 
\begin{align}
    \label{MIL_Formulation}
    Y_{i} &= 
    \begin{cases} 
        0, & \text{if } \sum_{b=1}^{B_i} y_{i,b} = 0, \quad y_{i,b} \in \{0,1\}, \\ 
        1, & \text{otherwise}. 
    \end{cases}
\end{align}
It can be interpreted as only if the labels of all instances are $0$ (negative), the final prediction is $0$ (negative). Otherwise, it would be labeled to $1$ (positive).

A common MIL framework consists of two trainable modules: feature extractor (FE) \(G_f\) and classifier \(G_c\). 
In this work, we use the pre-trained \(G_f\) (detailed in Section~\ref{sub:Implementation}) to embed \(x_{i,b}\) into a \(D\)-dimensional feature vector \(v_{i,b}\in\mathbb{R}^{1 \times D}\). 
The \(G_c\) is used to predict the class probability of \(X_{i}\), defined as \(P(\hat{Y}_{i}) = G_c(\{v_{i,b}\}_{b=1}^{B_i})\). Due to the variability of ${B_i}$, various issues arise in MIL approaches that use the same program for all instances $\{x_{i,b}\}_{b=1}^{B_i}$.

\subsection{Overview for PAMIL}%
\label{sub:PAMIL_Overview}
Figure~\ref{fig:figure2} outlines the framework of the proposed PAMIL. Our primary goal is to construct adaptive and reliable instance sampling and fusion schemes for WSI analysis by leveraging intrinsic relationships between past and current knowledge. 

Drawing inspiration from~\cite{Time_Decision}, we introduce a Markov Decision Process (MDP)~\cite{RLreview20} for PAMIL and enhance it using the RL algorithm. Key components of the MDP comprise: \{\textit{State}, \textit{Action}, \textit{Reward}\}. In this study, we define $\{v_{i,b}\}_{b=1}^{B_i}$ as state space \(\mathcal I_i^{B_t}\) and select randomly $v_{i}^{s_{t}}=\{v_{i,j}^{s_{t}}\}_{j=1}^{M}$ as the initial state at $t=1, t\in T$, where $M$ denotes the size of $v_{i}^{s_{t}}$. The $T=\frac{B_i}{M}$ is sampling step. To efficiently link MDP with the instance sampling, feature representation, and decision-making of MIL, unlike current methods~\cite{RLogist} to directly aggregate information from thousands of instances, we devise a dynamic policy-driven instance selection (DPIS) scheme to explore the intrinsic relation among the current instances, previous knowledge and remaining state space \(\mathcal I_i^{B_{t+1}}\). It encourages the network to select the most informative instances to facilitate the decision-making, detailed in Section~\ref{sec:DPIS}. Besides the instance sampling, we introduce a selection fusion feature representation (SFFR) module to aggregate information of the current features $v_{i}^{s_{t}}$ and past tokens $\{u_{i}^{s_{k}}\}_{k=1}^{t-1}$ for a stable and specific token $u_{i}^{s_{t}}$ representation. $u_{i}^{s_{t}}$ is passed to the DPIS and decision module for optimal sampling and robust prediction. Meanwhile, to enable DPIS to sample label-relevant and robust instances, instead of rewards from sparse WSI label support or teacher models~\cite{RLogist,FastMDP}, we introduce a self-guided feedback mechanism from the Transformer classification module (TCM), detailed in Section~\ref{sec:RPDMF}.

\subsection{Dynamic Policy Instance Selection Scheme}%
\begin{figure}[!t] \centering
\includegraphics[width=0.48\textwidth]{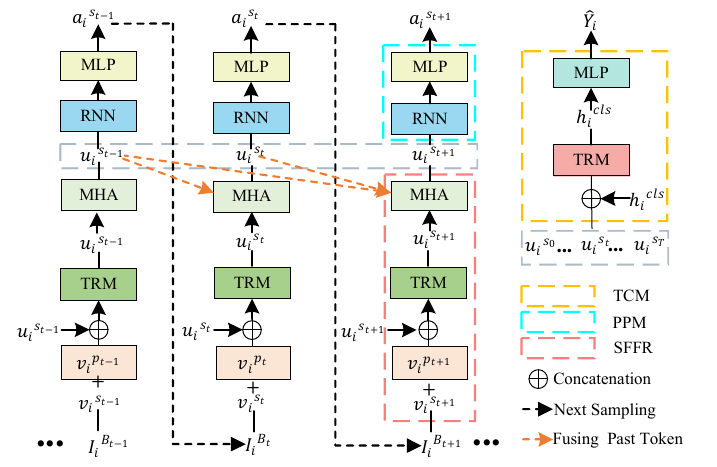}
    \caption{Illustration of the proposed \textbf{selection fusion feature representation (SFFR) module}, \textbf{proximal policy module} (PPM) and \textbf{Transformer-based classification module (TCM)} in PAMIL.}  \label{fig:figure3}
    \vspace{-5mm}
\end{figure}

\label{sec:DPIS}
DPIS aims to select the informative samples from the remaining instances with the guidance of previous knowledge. Therefore, we first introduce an experience-based learning of proximal policy module (PPM) ${G}_p$. As shown in Figure~\ref{fig:figure3}, $u_{i}^{s_{t}}$ is input into a recurrent neural network (RNN) ${G}_p^{\text{RNN}}$~\cite{RNN} with captured temporal dependencies, followed by a multi-layer perceptron (MLP) ${G}_p^{\text{MLP}}$ to guide the next sampling. This process is expressed as  
\begin{align}
\label{eq:Action_probabilities}
P_i^t (a_{i}^{s_{t}} | u_{i}^{s_{t}}) &= {G}_p^{\text{MLP}}({G}_p^{\text{RNN}}(u_{i}^{s_{t}})),
\end{align}
Where $a_{i}^{s_{t}}$ is next sampling instances indexes. The primary goal is to improve the sampling process. By considering the connectivity and proximity between the current and the remaining instances, three different schemes are introduced to optimise instance selection, including greedy policy-based max similarity scheme (GMSS), greedy policy-based hybrid similarity scheme (GHSS) and policy-optimized linear interpolation instances scheme (LIIS). \textbf{More details for sampling strategies are included in the Supplementary.}

\subsection{Selection Fusion Feature Representation}%
\label{sec:SFFR}

To aggregate robust and decision-friendly representations, we design the SFFR $G_s$ module to comprehensively explore relationships across current and historical information. As shown in Figure~\ref{fig:figure3}, we first initialize a vector $v_{i}^{p_{t}}$ to learn spatial information in $v_{i}^{s_{t}}$. The Transformer module (TRM) $G_s^{\text{TRM}}$ with an initial token $u_{i}^{s_{t}}$ is used to extract both local and global representations from $v_{i}^{s_{t}}$. To alleviate the catastrophic forgetting of historical information while thoroughly exploring the global texture, SFFR is equipped with a multi-head attention (MHA) mechanism $G_s^{\text{MHA}}$. Unlike randomly initializing vectors as tokens, MHA utilizes $u_{i}^{s_{t}}$ as a token to combine past tokens $\{u_{i}^{s_{k}}\}_{k=1}^{t-1}$ for enriched current representation. In addition, to promote the generalization of $u_{i}^{s_{t}}$, SFFR employs a more stable and efficient Siamese (SIA) structure~\cite{CLsiamese} among $u_{i}^{s_{t-1}}$ and $u_{i}^{s_{t}}$. The procedures in SFFR are expressed as 
\begin{align}
\label{State_Design02}
 u_{i}^{s_{t}} &= G_s^{\text{MHA}}([G_s^{\text{TRM}}( [ u_{i}^{s_{t}} ; (v_{i}^{p_{t}}+v_{i}^{s_{t}})  ]) ; \{u_{i}^{s_{k}}\}_{k=1}^{t-1}]),
\end{align}
\vspace{-2em}
\begin{align}
\label{eq:SIA}
\mathcal L_{\text{SIA}}^i = \frac{1}{T} \sum_{t=1}^{T} \left[\frac{1}{2} \mathcal D(p_{i}^{s_t}, u_{i}^{s_{t-1}}) + \frac{1}{2} \mathcal D (p_{i}^{s_{t-1}}, u_{i}^{s_t})\right],
\end{align}
where $p_{i}^{s_t}$ and $p_{i}^{s_{t-1}}$ are the MLP output and $\mathcal D$ refers to the negative cosine similarity~\cite{CLsiamese}.

\subsection{Decision-Making and Feedback}
\label{sec:RPDMF}
\paragraph{Decision-Making.} 
Considering the tumor features may be diluted in long sequences, potentially leading to false negatives, we use a MLP ${G}_{\text{s}}^{\text{MLP}}$ to infer the category $\{\hat{y}_{i,t}\}_{t=1}^T$ of $\{u_{i}^{s_{t}}\}_{t=1}^T$. In addition, to enhance prediction performance, we construct the Transformer classification module (TCM) ${G}_c^{\text{TCM}}$, as shown in Figure~\ref{fig:figure3}. The TCM further fuses $\{u_{i}^{s_{t}} \}_{t=1}^{T}$ via an initialized class token $h_{i}^{cls}$ for robust high-level global representation and employs a MLP ${{G}_{c}}^{\text{MLP}}$ to predict the label ${\hat{Y}}_{i}$. For the WSI task, the class token $h_{i}^{cls}$ prediction is used to minimize false positives due to noise. The final decision-making $\hat{Y}'_{i}$ for $X_i$ is depicted as
\vspace{-0.5em}
\begin{equation}
\hat{Y}'_{i} = 
\begin{cases} 
\frac{\hat{y}_{i, \text{max}} + \text{avg}(\hat{y}_{i,{1:3}}) + \text{avg}(\hat{y}_{i,{1:5}}) + \hat{Y}_i}{4}, & \text{if } Y = 1, \\
\hat{Y}_i, & \text{if } Y = 0.
\end{cases}
\end{equation}
where \(\hat{y}_{i,\text{max}}\), \(\text{avg}(\hat{y}_{i, {1:3}})\) and \(\text{avg}(\hat{y}_{i,{1:5}})\) denote the top-\(1\) max, and averages of the top \(3\) and \(5\) in \(\{\hat{y}_{i,t}\}_{t=1}^T\), respectively.

\paragraph{Reward and Punishment.} 
Although utilizing ${\hat{Y}}_{i}$ and $\{\hat{y}_{i,t}\}_{t=1}^T$ with label ${Y}_{i}$ consistency as rewards is straightforward, limited WSI-label ${Y}_{i}$ provides sparse guidance and approximate label prediction of $\hat{y}_{i,t}$ introduces bias. Moreover, while the most label-related instances are useful for decision-making in the inference stage, they are not conducive to training a well-generalized model.

Based on this analysis, we first introduce \(h_{i}^{cls}\) as target and $\{u_{i}^{s_{t}} \}_{t=1}^{T}$ as negative samples. The cosine similarity between \(h_{i}^{cls}\) and $\{u_{i}^{s_{t}} \}_{t=1}^{T}$ acts as the penalty term $r_i^p=\{r_{i,t}^p\}_{t=1}^T$. Meanwhile, we integrate prediction-labeling consistency to design the reward \(r_i^*\). These strategies guide the DPIS to select instances that aid in accuracy prediction and enhance stability. The feedback \(R_i\) is described by
\vspace{-1em}
\begin{align}
    \label{eq:reward_mechanism}
    R_{i} &= 
    \begin{cases} 
        r_i^* - r_i^p, & \text{if } \hat{Y}_i = Y_i, \\ 
        0 - r_i^p, & \text{otherwise}.
    \end{cases}
\end{align}

\subsection{Optimization Method}
\label{sec:loss}
\paragraph{Instance Sampling Scheme Optimization.}
\label{sec:loss_DPIS}
It is known that the proximal policy optimization (PPO)~\cite{PPO} algorithm is efficient in high-dimensional spaces and widely used in RL tasks~\cite{MuRCL,RLreview20}. We employ PPO to optimize sampling strategies combined with Eq.~\ref{eq:reward_mechanism}, focusing on relevant instances and ensuring robust bag representations.

\paragraph{Feature Representation and Decision-Making Optimization.}
\label{sec:loss_SFFR_TCG} 
To ensure the co-evolution of feature representation and decision-making, we design three losses to optimize both the SFFR and TCM (SFTC): the WSI probability ${\hat{Y}}_{i}$ cross-entropy loss (WSL) $\mathcal L_{\text{WSL}}^i$, the sub-bag token probability $\hat{y}_{i,t}$ cross-entropy loss (STL) $\mathcal L_{\text{STL}}^i$, and the $u_{i}^{s_{t-1}}$ to $u_{i}^{s_{t}}$ SIA contrastive loss~\cite{CLsiamese} $\mathcal L_{\text{SIA}}^i$ (Eq.~\ref{eq:SIA}). The losses are expressed as
\begin{align}
    \label{loss:WSI}
\mathcal L_{\text{WSL}}^i = - [Y_i\log \hat{Y_i} + (1 - Y_i) \log (1 - \hat{Y_i})],
\end{align}
\vspace{-2.5em}
\begin{align}
 \mathcal L_{\text{STL}}^i = - \frac{1}{T} \sum_{t=1}^{T} [Y_{i} \log \hat{y}_{i,t} + (1 - Y_{i}) \log(1 - \hat{y}_{i,t})].
\end{align}
The total loss function is given by
\vspace{-1em}
\begin{align}
    \mathcal L_{\text{SFTC}} = \frac{1}{N}\sum _{i=1}^{N} [\mathcal L_{\text{WSL}}^i + \lambda_{\text{STL}} \cdot \mathcal L_{\text{STL}}^i +\lambda_{\text{SIA}} \cdot \mathcal L_{\text{SIA}}^i ],
\end{align}
where $\lambda_{\text{STL}}$ and $\lambda_{\text{SIA}}$ 
are adjusted from initial to final value via integrating epoch and cosine curve~\cite{CosineAnnealing}. For more details on the optimization methods, please refer to the \textbf{Supplementary}.

\section{Experiments}
\begin{table*}[t] \centering
    \newcommand{\Frst}[1]{\textcolor{red}{\textbf{#1}}}
    \newcommand{\Scnd}[1]{\textcolor{blue}{\textbf{#1}}}
    \caption{\textbf{Quantitative comparison of our results and pseudo-bags-level methods on CAMELYON16 and TCGA Lung datasets.} Results from~\cite{ReviewMILWSI1, ABMIL, CLAM, CVPR21DSMIL, Transmil, RNNMIL} are derived from DTFD~\cite{DTFDMIL}, with others~\cite{22pseudo-labelDgmil, MHIM, IAT, DTFDMIL} from their published papers. The numbers in {\color{red}\textbf{red}}, 
 {\color{blue}\textbf{blue}} and \underline{underline} indicate the best, second best, and previous best performance, respectively.}
    \label{tab:Table1_PAMIL_Comp_SOTA}
    \newcommand{\normal}{\boldsymbol{n}}
\newcommand{\point}{\boldsymbol{x}}

\makebox[\textwidth]{\small} 
\resizebox{1.0\textwidth}{!}{
    %\tiny
 \scriptsize %\small
\begin{tabular}{l|*{3}{c}|*{3}{c}}
    \hline
    {\multirow{2}{*}{Methods}}& \multicolumn{3}{c|}{CAMELYON16} & \multicolumn{3}{c}{TCGA Lung} \\
    \multicolumn{1}{c|}{} & Accuracy & F1 & AUC & Accuracy & F1 & AUC \\
    \hline
    \multicolumn{1}{c|}{MeanPooling~\cite{ReviewMILWSI1}} & 0.626$\pm$0.008 & 0.355$\pm$0.007 & 0.528$\pm$0.008 & 0.833$\pm$0.011 & 0.809$\pm$0.012 & 0.901$\pm$0.012 \\
    \multicolumn{1}{c|}{MaxPooling~\cite{ReviewMILWSI1}} & 0.826$\pm$0.023 & 0.754$\pm$0.048 & 0.854$\pm$0.030 & 0.846$\pm$0.029 & 0.833$\pm$0.027 & 0.901$\pm$0.033 \\
    \multicolumn{1}{c|}{ABMIL~\cite{ABMIL}} & 0.845$\pm$0.005 & 0.780$\pm$0.009 & 0.854$\pm$0.005 & 0.869$\pm$0.032 & 0.866$\pm$0.021 & 0.941$\pm$0.028 \\
   \multicolumn{1}{c|}{RNNMIL~\cite{RNNMIL}}
    & 0.844$\pm$0.021 & 0.798$\pm$0.006 & 0.875$\pm$0.002 & 0.845$\pm$0.024 & 0.831$\pm$0.023 & 0.894$\pm$0.025 \\
    \multicolumn{1}{c|}{DSMIL~\cite{CVPR21DSMIL}} & 0.856$\pm$0.010 & 0.815$\pm$0.014 & 0.899$\pm$0.007 & 0.888$\pm$0.013 & 0.876$\pm$0.011 & 0.939$\pm$0.019 \\
    \multicolumn{1}{c|}{CLAM-MB~\cite{CLAM}} & 0.823$\pm$0.022 & 0.774$\pm$0.017 & 0.878$\pm$0.013 & 0.878$\pm$0.043 & 0.874$\pm$0.028 & 0.949$\pm$0.019 \\
    \multicolumn{1}{c|}{CLAM-SB~\cite{CLAM}} & 0.837$\pm$0.023 & 0.775$\pm$0.016 & 0.871$\pm$0.012 & 0.875$\pm$0.041 & 0.864$\pm$0.043 & 0.944$\pm$0.023 \\
    \multicolumn{1}{c|}{TransMIL~\cite{Transmil}} & 0.858$\pm$0.008 & 0.797$\pm$0.017 & 0.906$\pm$0.025 & 0.883$\pm$0.022 & 0.876$\pm$0.021 & 0.949$\pm$0.013 \\
    \multicolumn{1}{c|}{DTFD (AFS)~\cite{DTFDMIL}} & 0.908$\pm$0.013 & 0.882$\pm$0.017 & 0.946$\pm$0.004 & 0.891$\pm$0.033 & 0.883$\pm$0.025 & 0.951$\pm$0.022 \\
    \multicolumn{1}{c|}{DTFD (MaxMinS)~\cite{DTFDMIL}} & 0.899$\pm$0.010 & 0.865$\pm$0.014 & 0.941$\pm$0.003 & 0.894$\pm$0.033 & 0.891$\pm$0.027 & 0.961$\pm$0.021 \\
    \multicolumn{1}{c|}{DGMIL~\cite{22pseudo-labelDgmil}} & 0.802$\pm$ --- & --- & 0.837$\pm$ --- & 0.920$\pm$ --- & --- & \underline{0.970$\pm$ ---} \\
    \multicolumn{1}{c|}{IAT~\cite{IAT}} & 0.899$\pm$0.001 & 0.874$\pm$0.009 & 0.946$\pm$0.006 & \underline{0.921$\pm$0.008} & 0.849$\pm$0.015 & 0.849$\pm$0.016 \\
    \multicolumn{1}{c|}{MHIM (TransMIL)~\cite{MHIM}} & 0.920$\pm$0.890 & 0.901$\pm$1.080 & \underline{0.965$\pm$0.480} & 0.900$\pm$2.590 & \underline{0.897$\pm$2.630} & 0.949$\pm$2.170 \\
    \multicolumn{1}{c|}{MHIM (DSMIL)~\cite{MHIM}} & \underline{0.925$\pm$0.350} & \underline{0.908$\pm$0.730} & 0.965$\pm$0.650 & 0.898$\pm$3.370 & 0.897$\pm$2.920 & 0.955$\pm$1.740 \\
     \hline
   \multicolumn{1}{c|}{DPIS-GMSS} & \Frst{0.963$\pm$0.033} & \Frst{0.948$\pm$0.048} & \Scnd{0.971$\pm$0.044} & 0.953$\pm$0.002 & 0.949$\pm$0.003 & 0.988$\pm$0.003 \\
   \multicolumn{1}{c|}{DPIS-GHSS} & 0.942$\pm$0.012 & 0.922$\pm$0.017 & 0.970$\pm$0.017 & \Scnd{0.962$\pm$0.003} & \Scnd{0.958$\pm$0.004} & \Scnd{0.990$\pm$0.003} \\
   \multicolumn{1}{c|}{DPIS-LIIS} & \Scnd{0.961$\pm$0.006} & \Scnd{0.947$\pm$0.009} & \Frst{0.977$\pm$0.026} & \Frst{0.965$\pm$0.007} & \Frst{0.961$\pm$0.008} & \Frst{0.994$\pm$0.002} \\
    \hline
\end{tabular}
}

\end{table*}
\begin{table*}[t] \centering
    \newcommand{\Frst}[1]{\textcolor{red}{\textbf{#1}}}
    \newcommand{\Scnd}[1]{\textcolor{blue}{\textbf{#1}}}
    \vspace{-3mm} 
    \caption{\textbf{Quantitative comparison of our results and pseudo-bags schemes on CAMELYON16 and TCGA Lung datasets.} $^\S$ indicates $512$ instances per group, selected either randomly or by positional sorting. $^\sharp$ denotes k-Means or random grouping of each bag into $10$ groups. The results come from  the class token $h_{i}^{cls}$ prediction.
    }  
    \label{tab:Table2_DPIS_Comp_Pseudo_Bag}
    \newcommand{\normal}{\boldsymbol{n}}
\newcommand{\point}{\boldsymbol{x}}
\makebox[\textwidth]{\small}

\resizebox{\textwidth}{!}{
 \scriptsize %\Large
\begin{tabular}{l|c|*{5}{c}|*{5}{c}}
    \hline
    \multicolumn{2}{c|}{\multirow{2}{*}{Methods}}  &\multicolumn{5}{c|}{CAMELYON16} & \multicolumn{5}{c}{TCGA Lung} \\
   \multicolumn{2}{c|}{}& Accuracy & Precision & Recall & F1 & AUC & Accuracy & Precision & Recall & F1 & AUC \\
    \hline
    \multicolumn{1}{c|}{Position Sorting $^\S$} & 512 & 0.861 & 0.878 & 0.735 & 0.800 & 0.942 & 0.947 & 0.934 & 0.950 & 0.942 & \Scnd{0.991} \\
    \multicolumn{1}{c|}{Random Sampling $^\S$} & 512 & 0.861 & 0.878 & 0.735 & 0.800 & 0.897 & 0.947 & 0.920 & 0.966 & 0.943 & 0.990 \\
    \multicolumn{1}{c|}{K-Means Grouping $^\sharp$} & 10 & 0.899 & 0.950 & 0.776 & 0.854 & 0.940 & 0.947 & 0.927 & 0.958 & 0.942 & 0.990 \\
    Random Grouping $^\sharp$ & 10 & 0.920 & 0.924 & \Scnd{0.861} & 0.891 & 0.926 & 0.951 & 0.914 & \Scnd{0.983} & 0.947 & 0.989 \\
    \hline
    \multicolumn{2}{c|}{DPIS-GMSS} & 0.915 & \Scnd{0.975} &  0.800 & 0.876 & 0.905 & \Scnd{0.954} &  0.921 & \Frst{0.983} & \Scnd{0.951} & 0.987 \\
    \multicolumn{2}{c|}{DPIS-GHSS} & \Scnd{0.923} & 0.933 & 0.857 & \Scnd{0.894} & \Frst{0.944} & \Frst{0.958} & \Scnd{0.950} & 0.958 & \Frst{0.954} & 0.989 \\
    \multicolumn{2}{c|}{DPIS-LIIS} & \Frst{0.954} & \Frst{0.978} & \Frst{0.898} & \Frst{0.936} & \Scnd{0.944} & 0.954 & \Frst{0.965} & 0.933 & 0.949 & \Frst{0.992} \\
    \hline
\end{tabular}
}

\end{table*}\vspace{-2mm}

\begin{figure*}[t] \centering
    \includegraphics[width=0.95\textwidth]{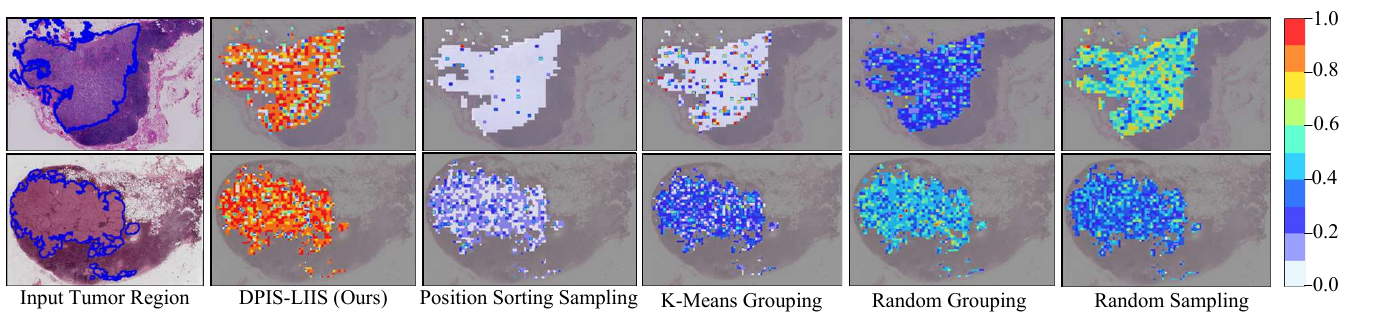}
    \caption{\textbf{Visual comparisons with pseudo-bags-level methods on CAMELYON16 dataset.} The blue outline indicates the tumor region. Attention score bar (0-1): indicates model focus on tumor instances, higher values show more attention, not positive probability.} \vspace{-4mm}
 \label{fig:figure4}
\end{figure*}

\label{sec:Experiments}
To validate our PAMIL, we conduct extensive experiments on the CAMELYON16~\cite{CAMELYON16} and TCGA Lung Cancer~\cite{TCGALung} datasets with mainstream MIL methods, involving bag-level methods (MeanPooling~\cite{ReviewMILWSI1}, MaxPooling~\cite{ReviewMILWSI1}, ABMIL~\cite{ABMIL}, RNNMIL~\cite{RNNMIL}, DSMIL~\cite{CVPR21DSMIL}, CLAM~\cite{CLAM}, TransMIL~\cite{Transmil}, IAT~\cite{IAT}, and MHIM~\cite{MHIM}) and pseudo-bags or pseudo-labels methods (DTFD~\cite{DTFDMIL} and DGMIL~\cite{22pseudo-labelDgmil}). Subsequently, we evaluate common grouping methods within our SFFR and TCM. Meanwhile, ablation studies are conducted to verify individual effects of basic components on the final performance. %

\subsection{Datasets and Metrics}
\label{sub:Dataset_and_Metrics}
\paragraph{CAMELYON16 Dataset~\cite{CAMELYON16}.} The dataset consists of $270$ training WSIs ($159$ normal, $111$ tumor) and $129$ testing WSIs for breast cancer lymph nodes. Following TransMIL~\cite{Transmil}, we apply CLAM~\cite{CLAM} to identify tissues on WSIs and obtain non-overlapping $256\times256$ instances at $20\times$ magnification. 

\paragraph{TCGA Lung Cancer Dataset~\cite{TCGALung}.} This database contains two cancer sub-categories: Lung Adenocarcinoma (LUAD) and Lung Squamous Cell Carcinoma (LUSC). It provides 541 LUAD slides from 478 cases and 512 LUSC slides from 478 distinct cases. We adopt the same pre-processing as DSMIL~\cite{CVPR21DSMIL} for 1046 WSIs. Each WSI is then segmented into non-overlapping patches of $224\times224$ at $20\times$ magnification.
\paragraph{Evaluation Metrics.}
The commonly used metrics, like Area Under Curve (AUC), accuracy, and F1 score (F1) are introduced for evaluation. Precision and recall are also considered, with a threshold of $0.5$. We adopt the same settings as~\cite{DTFDMIL,Transmil,MMIL,MHIM} for a fair comparison due to discrepancies of dataset splits. The CAMELYON16 official training set is further randomly divided into training and validation sets at $9:1$. The TCGA Lung is randomly split into training, validation, and testing sets with ratios of $65:10:25$. 
Detailed data processing is in the \textbf{Supplementary}.
\subsection{Implementation Details}%
\label{sub:Implementation}
For the CAMELYON16 dataset, following~\cite{CLAM,Transmil,DTFDMIL,MHIM}, we extract $1024$-dimensional feature vectors from each instance via a pre-trained ResNet50~\cite{ResNet18} with ImageNet~\cite{Imagenet}. For the TCGA Lung dataset, we employ the SimCLR~\cite{SimCLR} with a ResNet18~\cite{ResNet18} encoder, 
to obtain $512$-dimension feature vectors from each patch. AdaMax optimizer~\cite{Adam} with a weight decay of $1e-5$ and the initial learning rate of $1e-4$ are used. With the above settings, we train our PAMIL with $300$ epochs with batch size $1$ on one NVIDIA 2080Ti GPU. 
\subsection{Comparison with State-of-the-Art}%
\label{sub:Comparisons with State-of-the-Art Methods}
Quantitative results are presented in Table~\ref{tab:Table1_PAMIL_Comp_SOTA}. 
As expected, our method achieves the best scores across all metrics and datasets, surpassing the SOTA MHIM~\cite{MHIM} by 3.8\% in terms of accuracy on the CAMELYON16 dataset and over IAT~\cite{IAT} by 4.4\% with accuracy on the TCGA lung over. It is noted that the competitors obtain impressive performance on the TCGA lung dataset with high consistency, as the TCGA lung dataset consists of more than 80\% tumor areas~\cite{Transmil}. Since positive WSIs on the CAMELYON16 dataset occupies only small portions of tumor~\cite{Transmil, CVPR21DSMIL}, it is observed that the bag-level methods struggle to capture positive features to drive a credible decision~\cite{ReviewMILWSI1,ABMIL,RNNMIL,CLAM}. Exploring the global response with attention-based and Transformer-based methods~\cite{Transmil,IAT,CVPR21DSMIL} achieve better performance by producing discriminant representations across numerous samples. However, it is still behind the pseudo-bags~\cite{DTFDMIL}. In particular, MHIM~\cite{MHIM} improves bag generalization features by masking key instances with well accuracy, yet it falls 3.8\% behind our method. The main reason is that these methods barely explore the intrinsic relations between the historical knowledge and current instance during sampling and feature representation. It greatly facilitates robust and generalized performance. In addition, DGMIL~\cite{22pseudo-labelDgmil} employs K-Means grouping introducing a prior bias, resulting in poor performance. These results further demonstrate the effectiveness of our policy-driven instance selection and election fusion feature representation schemes. 

\subsection{Comparison with Pseudo-Bags Schemes} 
\paragraph{Quantitative Comparison.} 
To investigate the effects of sampling strategies, Table~\ref{tab:Table2_DPIS_Comp_Pseudo_Bag} tabulates the quantitative comparisons against different pseudo-bags schemes, including the commonly used position sorting, random sampling and group, as well as our proposed greedy policy-based max similarity scheme (GMSS), greedy policy-based hybrid similarity scheme (GHSS) and policy-optimized linear interpolation instances scheme (LIIS). Notably, most of the strategies demonstrate impressive performance on the TCGA lung dataset. Our method still performs favorably on the challenging CAMELYON16 dataset. Specifically, our LIIS scheme improves accuracy by 9.3\% and 3.4\% against random sampling and random grouping, respectively. Although position sorting and K-Means schemes consider the spatial and features relationships, they struggle to exploit contextual prior, leading to undesired accuracy, respectively lower than LIIS and GHSS by 9.3\%, 5.5\%, 6.2\% and 2.4\%. We speculate that these visible improvements of scores benefit from the elaborate sampling schemes, which encourage the network to aggregate the knowledge from historical instance representation according to the similarity. It is crucial to guide the instance sampling and robust decision-making.

\paragraph{Qualitative Comparison and Interpretability.}
\label{sub:Quantitative}
Figure~\ref{fig:figure4} visualizes the attention scores from the last self-attention layer in TCM, demonstrating that our scheme can encourage model to focus on tumor regions. It is barely explored in pseudo-bags methods. Furthermore, when grouping based on prior context and position, biases in the data lead the model to focus on instances that are not label-related. This evidence not only supports previous analyses but also suggests that our scheme facilitates more accurate learning of bag representations that are closely associated with the labels.
\begin{table}[!t]\centering
\vspace{-2mm} 
    \caption{\textbf{Quantitative results of DPIS and SFFR (DPSF) scheme for four MIL methods on CAMELYON16 dataset.} The \textbf{bold} indicates an improvement over previous results. The results are based on the class token $h_{i}^{cls}$ prediction.}  
    \label{tab:Table3_DPSF_Adapt_MIL}
    \newcommand{\normal}{\boldsymbol{n}}
\newcommand{\point}{\boldsymbol{x}}

\renewcommand{\arraystretch}{1.2} % Increase the row height
\resizebox{\columnwidth}{!}{
\scriptsize
\setlength{\tabcolsep}{2pt}
% \begin{tabular}{l*{5}{c}}
\begin{tabular}{c|cccc}
    \hline
    Methods & Accuracy   & F1 & AUC \\
    \hline
    \text{MaxPooling}~\cite{ReviewMILWSI1}  & 0.826 & 0.754 &  0.854\\
    DPSF-MaxPooling & 0.930 \textbf{(+10.4\%)} & 0.901\textbf{(+14.7\%)} & 0.938\textbf{(+8.4\%)}\\
    \hline
    \text{MeanPooling}~\cite{ReviewMILWSI1} & 0.626 & 0.355 & 0.528 \\
     DPSF-MeanPooling & 0.876\textbf{(+25.0\%) } & 0.830\textbf{(+47.5\%)} & 0.916\textbf{(+38.0\%)}\\
    \hline
    \text{DSMIL}~\cite{CVPR21DSMIL} &0.856 &0.815 &0.899 \\
    DPSF-DSMIL & 0.907\textbf{(+5.1\%)}   & 0.882\textbf{(+6.7\%) } & 0.935\textbf{(+4.6\%)} \\
    \hline
    \text{TransMIL}~\cite{Transmil} & 0.858 & 0.797 & 0.906 \\
     DPSF-TransMIL &0.884  \textbf{(+2.6\%)}  &  0.845  \textbf{(+4.8\%)}   & 0.926 \textbf{(+2.0 \%)} \\
     \hline
\end{tabular}
}

      \vspace{-3mm} 
\label{tab:DPSF}
\end{table}

\subsection{Adaptability for MIL Method}
Due to the considerable versatility and scalability of our proposed DPIS and SFFR (unified into the DPSF module), we equip four representative MIL frameworks with the DPSF module and evaluate them on the CAMELYON16 dataset. The quantitative results, tabulated in Table~\ref{tab:DPSF}, demonstrate significant improvements over existing results. These findings validate the flexibility of DPIS and confirm that SFFR is proficient in generating discriminative representations to boost prediction. Intuitively, DPSF-MaxPooling selects the most salient tokens from $\{u_{i}^{s_t}\}_{t=1}^{T}$ for prediction and obtains optimal results compared to DPSF-DSMIL and DPSF-TransMIL. We attribute this to SFFR providing the token with highly label-consistent features. DPSF-MeanPooling shows a 38\% improvement in AUC over standard MeanPooling. However, due to the uncertain representation from input instances, DPSF-MeanPooling generates bag representations by averaging $\{u_{i}^{s_t}\}_{t=1}^{T}$ inferior attention fusion. 
\begin{figure}[t] \centering
\includegraphics[width=0.45\textwidth]{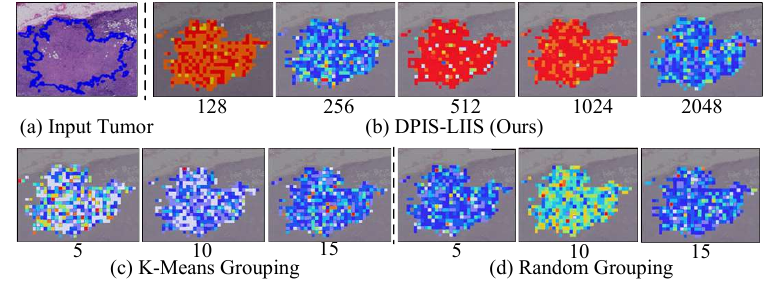}
    \caption{\textbf{Visual effects of pseudo-bag sizes and grouping schemes on CAMELYON16 dataset.}} \vspace{-3mm} \label{fig:figure5}
\end{figure}
\begin{figure}[!t] \centering
\includegraphics[width=0.45\textwidth]{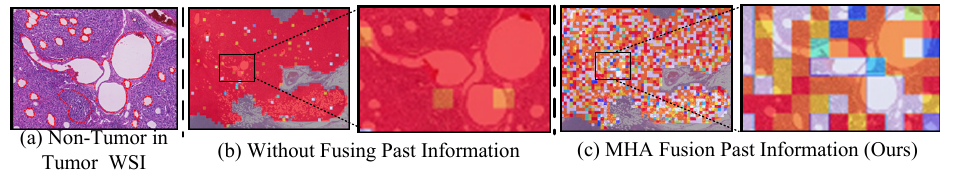}
    \caption{\textbf{Visual effects of selection fusion feature representation on CAMELYON16 dataset.} 
    (a) Red contours indicate non-tumor cells in the tumor WSI. (b) and (c) Black boxes indicate the attention of different fusion mechanisms to non-tumor instances.
    } \vspace{-4mm} \label{fig:figure6}
\end{figure}

\subsection{Ablation Study}%
To validate the PAMIL and analyze the contributions of individual components, comprehensive studies are conducted using the LIIS scheme, specifically sampling from the CAMELYON16 dataset.

\paragraph{Effects of Pseudo-bag Size and Grouping Scheme.}
Figure~\ref{fig:figure5} and Table~\ref{tab:Table4_Ablation_study}(a) display visual and numerical comparisons that reveal a small pseudo-bag lack the detail needed for accurate tumor-normal feature discrimination, resulting in higher false-negative rates. On the other hand, a large pseudo-bag may obscure sparse tumor instance correlations, leading to more false positives. Comparatively, fixed grouping schemes exhibit lower accuracy and less attention to tumor instances than our proposed DPIS strategy. This affirms the shortcomings of MIL methods stated in Section~\ref{sec:intro} and emphasizes the need for dynamic instance sampling in WSI analysis.

\begin{table}[t]\centering
\newcommand{\Frst}[1]{\textcolor{red}{\textbf{#1}}}
\newcommand{\Scnd}[1]{\textcolor{blue}{\textbf{#1}}}
\vspace{-3mm} 
    \caption{\textbf{Ablation study on the CAMELYON16 dataset}. The baseline uses randomly selected 512 instances to form a pseudo-bag for training SFFR and TCM under the loss defined in Eq~\ref{loss:WSI}. The results are based on the class token $h_{i}^{cls}$ prediction.}
    \label{tab:Table4_Ablation_study}
    % \makebox[\textwidth]{\small (a) Results on dataset A.} 
\caption*{(a) \textbf{Effects of pseudo-bag size and grouping scheme}. ``Mix'' and ``Max'' indicate pseudo-bag counts from the smallest and largest WSIs. $M$ denotes the instance count sampled by DPIS at time $t$, while ``KM'' and ``RG'' represent K-Means or random grouping of each bag into 5, 10, or 15 groups. We set $ M=512$ in our DPIS. }
\resizebox{\columnwidth}{!}{
 \scriptsize

\begin{tabular}{c|c|cccc|cc}
    \hline
    \multicolumn{2}{c|}{Methods}  & Accuracy & Precision & Recall & F1 & Mix & Max \\
    \hline
   \multirow{4}{*}{$M$}& 128 & 0.923 & 0.954 & 0.837 & 0.891 & 11 & 437 \\
    & 256 & 0.923 & \Scnd{0.976} & 0.816 & 0.889 & 6 & 219 \\
    & 512 & \Frst{0.954} & \Frst{0.978} & \Frst{0.898} & \Frst{0.936} & 3 & 110 \\
    & 1024 & \Scnd{0.930} & 0.935 & \Scnd{0.878} & \Scnd{0.905} & 2 & 55 \\
    & 2048 & 0.907 & 0.894 & 0.857 & 0.875 & 1 & 28 \\
    \hline
    \multirow{4}{*}{KM}& 5 & 0.907 & 0.974 & 0.776 & 0.864 & 5 & 5 \\
    & 10 & 0.899 & 0.950 & 0.776 & 0.854 & 10 & 10 \\
   &  15 & 0.915 & 0.913 & 0.857 & 0.884 & 15 & 15 \\
   \hline
    \multirow{4}{*}{RG} & 5 & 0.884 & 0.925 & 0.755 & 0.831 & 5 & 5 \\
    & 10 & 0.920 & 0.924 & 0.861 & 0.891 & 10 & 10 \\
   & 15 & 0.907 & 0.894 & 0.857 & 0.875 & 15 & 15 \\
    \hline
   \multicolumn{2}{c|}{Baseline} & 0.861 & 0.878 & 0.735 & 0.800 & 3 & 110 \\
    \hline
\end{tabular}
}
\vspace{0.1em}
\caption*{(b) \textbf{Effects of selection fusion feature representation.}}
\resizebox{\columnwidth}{!}{ % Resize table to fit the column width
\scriptsize
\begin{tabular}{cc|cccc}
    \hline
    \text{Attention} &   \text{MHA} & Accuracy & Precision & Recall & F1   \\
    \hline
    \ding{55}&\ding{55} & 0.938 & 0.956 & 0.878 & 0.915  \\
    \ding{51}&\ding{55}  & \Scnd{0.946} & \Scnd{0.977} & \Scnd{0.878} & \Scnd{0.925}  \\
     \ding{55}&\ding{51} & \Frst{0.954} & \Frst{0.978} & \Frst{0.898} & \Frst{0.936}  \\
    \hline
\end{tabular}
}
\vspace{0.1em}
\caption*{(c) \textbf{Effects of optimizing strategy.}}
\resizebox{\columnwidth}{!}{ % Resize table to fit the column width
\scriptsize
\begin{tabular}{ccc|cccc}
    \hline
    \(\mathcal L_{\text{WSL}}\) &\(\mathcal L_{\text{SIA}}\) & \(\mathcal L_{\text{STL}}\)& Accuracy & Precision & Recall & F1  \\
    \hline
    \ding{51} &\ding{55} &\ding{55} & 0.915 & 0.896 & 0.878 & 0.887 \\
    \ding{51} &\ding{51} &\ding{55} &\Scnd{0.930} & \Scnd{0.935} & \Scnd{0.878} & \Scnd{0.905} \\
    \ding{51} &\ding{55} &\ding{51} &0.915 & 0.913 & 0.857  & 0.884 \\
    \ding{51}&\ding{51} &\ding{51} &\Frst{0.954} & \Frst{0.978}	& \Frst{0.898} & \Frst{0.936} \\
   \hline
\end{tabular}
}

\renewcommand{\arraystretch}{1.1} % Increase the row height
\vspace{0.1em}
\caption*{(d) \textbf{Effects of reward and punishment}.}
\resizebox{\columnwidth}{!}{ % Resize table to fit the column width
\scriptsize

{
\begin{tabular}{ccc|cccc}
    \hline
    Model&$r_i^p$ & $r_i^*$ &  Accuracy  & Precision & Recall & F1 \\
    \hline
     \textit{w/o} $R_i$&\ding{55} &\ding{55}  & 0.930 & 0.917 & \Scnd{0.898} & 0.907 \\
     \textit{w} $r_i^*$ &\ding{55} &\ding{51} & 0.915 & 0.932 & 0.837 & 0.882 \\
     \textit{w} $r_i^p$ &\ding{51}&\ding{55}   & \Scnd{0.938} & \Scnd{0.956} & 0.878  & \Scnd{0.915} \\
     \textit{w} $R_i$ &\ding{51} & \ding{51} & \Frst{0.954} & \Frst{0.978} & \Frst{0.898} & \Frst{0.936} \\
    \hline
\end{tabular}}
}

    \vspace{-2mm} 
\end{table}

\paragraph{Effects of Selection Fusion Feature Representation.}
To demonstrate that incorporating features between past and current sampling instances facilitates robust and specific representation, we constructed three fusion experiments for $G_s^{\text{MHA}}$ in the SFFR. The visual and quantitative results are shown in Figure~\ref{fig:figure6} and Table~\ref{tab:Table4_Ablation_study}(b), showing that incorporating historical tokens enhances model focus on label-related features. The MHA outperforms softmax attention, further demonstrating that a fully fused feature representation yields a more precise bag representation.

\paragraph{Effects of Optimizing Strategy.}
Table~\ref{tab:Table4_Ablation_study}(c) demonstrates that our sampling strategy improves accuracy by 5.4\% compared to the baseline, using the same loss Eq~\ref{loss:WSI}. This indicates the advantage of distance-related sampling for better bag representation. Additionally, the contrast loss in adjacent pseudo-bags further enhances features, resulting in improved inference accuracy. However, it is important to note that highly sensitive pseudo-labels can introduce biases. To address this, a hybrid loss function produces optimal results by enhancing the differentiation between tumor and normal features in imbalanced WSI datasets.

\paragraph{Effects of Reward and Punishment.}
The reward and punishment mechanisms are crucial in the DPIS scheme. Integrating penalty and reward terms, as shown in Table~\ref{tab:Table4_Ablation_study}(d), significantly enhances model accuracy by promoting attention to robust and common features. Instead of using $r_i^{*}$, penalizing based on the similarity values $h_{i}^{cls}$ and $u_{i}^{s_{t}}$ can effectively reduce the false-positive rate by correcting biases in pseudo-bags and diversifying bag representation. More in-depth analyses of reward and punishment system are inluded in the \textbf{Supplementary}.

\section{Conclusion}%
\label{sec:Conclusion}
In this study, we propose a new paradigm, termed the dynamic policy-driven adaptive multi-instance learning framework (PAMIL) for WSI. The main focus is on investigating the interactions between instance sampling, feature representation, and decision-making. To enhance the accuracy of inference, we utilize past predictions and sampled instances to guide the selection of optimal features in the next step. To achieve this, we introduce a DPIS scheme that effectively incorporates current and previous rich information, along with reward-penalty mechanisms to optimize sampling. Additionally, the SFFR method is employed to filter pseudo-bags and improve bag representation for more robust predictions. Extensive experiments have been conducted, demonstrating the impressive performance of our proposed PAMIL and validating the effectiveness of exploring interrelationships. However, it is worth noting that the complex structural information of WSI is not fully exploited by the distance guide strategies. A potential improvement would involve refining the representation of the state token, which could be beneficial for sampling.

\noindent \textbf{Acknowledgement.}
This research was funded by the National Science and Technology Major Project (2021ZD0110901).

% \clearpage
{\small
\bibliographystyle{ieee_fullname}
\bibliography{ref}
}

\end{document}